%% file: main.tex
\documentclass[10pt,twocolumn,letterpaper]{article}

\usepackage[pagenumbers]{cvpr} 

\input{preamble}
\definecolor{cvprblue}{rgb}{0.21,0.49,0.74}
\usepackage[pagebackref,breaklinks,colorlinks,allcolors=cvprblue]{hyperref}

\title{CliPPER: Contextual Video-Language Pretraining on Long-form Intraoperative Surgical Procedures for Event Recognition}

\author{
Florian Philipp Stilz$^{1,2,3}$ \quad Vinkle Srivastav$^{1,2}$ \quad Nassir Navab$^{3}$ \quad Nicolas Padoy$^{1,2}$\\
$^{1}$University of Strasbourg, CNRS, INSERM, ICube, UMR7357, France\\
$^{2}$IHU Strasbourg, France\\
$^{3}$Technical University of Munich, Germany\\
{\tt\small florian.stilz@tum.de}
}

\begin{document}
\twocolumn[{
\renewcommand\twocolumn[1][]{#1}
\maketitle
\begin{center}
    \vspace{-8mm}
    \includegraphics[width=0.98\linewidth]{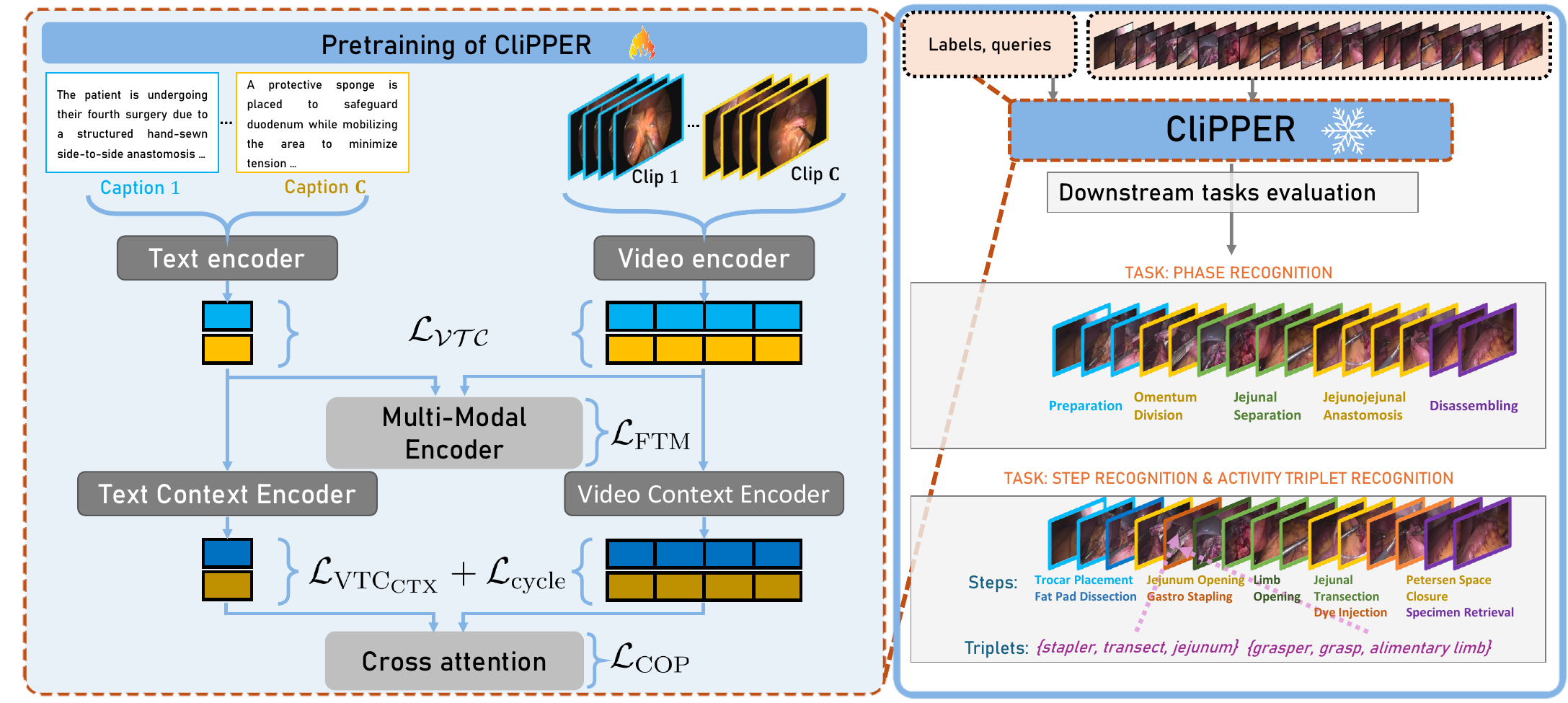}
    \captionof{figure}{\textbf{Overall CliPPER Framework:} \textit{CliPPER} is a powerful pretraining framework for surgical video–language understanding, designed to capture long-form and contextual relationships. It leverages a suite of novel objectives: including context-aware contrastive learning ($\mathcal{L}_{VTC_{CTX}}$) and Cycle-Consistency Alignment loss ($\mathcal{L}_{cycle}$), Frame-Text Matching ($\mathcal{L}_{FTM}$) for pinpointing relevant frames in long procedures, and Clip Order Prediction ($\mathcal{L}_{COP}$) for temporal reasoning. CliPPER demonstrates strong performance across a wide range of complex surgical downstream tasks - from high-level phase recognition to fine-grained activity triplet identification - without requiring any additional training. 
    }
    \label{fig:teaser}
\end{center}
}]

\input{sec/0_abstract}    
\input{sec/1_intro}
\input{sec/2_related}
\input{sec/3_method}
\input{sec/4_experiments}
\input{sec/5_conclusion}
\input{sec/6_acknowledgement}
{
    \small
    \bibliographystyle{ieeenat_fullname}
    \bibliography{main}
}
\input{sec/7_appendix}


\end{document}

%% file: sec/0_abstract.tex
\begin{abstract}
Video-language foundation models have proven to be highly effective in zero-shot applications across a wide range of tasks. A particularly challenging area is the intraoperative surgical procedure domain, where labeled data is scarce, and precise temporal understanding is often required for complex downstream tasks. To address this challenge, we introduce \textbf{CliPPER} (\textbf{C}ontextua\textbf{l} V\textbf{i}deo-Language \textbf{P}retraining on Long-form Intraoperative Surgical \textbf{P}rocedures for \textbf{E}vent \textbf{R}ecognition), a novel video-language pretraining framework trained on surgical lecture videos. Our method is designed for fine-grained temporal video-text recognition and introduces several novel pretraining strategies to improve multimodal alignment in long-form surgical videos. Specifically, we propose \textbf{Contextual Video–Text Contrastive Learning ($\mathbf{VTC_{CTX}}$)} and \textbf{Clip Order Prediction (COP)} pretraining objectives, both of which leverage temporal and contextual dependencies to enhance local video understanding. In addition, we incorporate a \textbf{Cycle-Consistency Alignment} over video–text matches within the same surgical video to enforce bidirectional consistency and improve overall representation coherence. Moreover, we introduce a more refined alignment loss, \textbf{Frame-Text Matching (FTM)}, to improve the alignment between video frames and text. As a result, our model establishes a new state-of-the-art across multiple \textbf{public surgical benchmarks}, including zero-shot recognition of phases, steps, instruments, and triplets. The source code and pretraining captions can be found at
\href{https://github.com/CAMMA-public/CliPPER}{https://github.com/CAMMA-public/CliPPER}.
\end{abstract}

%% file: sec/1_intro.tex
\section{Introduction}
\label{sec:intro}

Surgical care presents a profound paradox. Despite carrying the word ``care'' in its name, it remains the third leading cause of patient mortality and morbidity worldwide, trailing only heart disease and stroke~\cite{nepogodiev2019global}. This burden largely stems from the intraoperative adverse events, ranging from major complications to technical missteps. Surprisingly, around $41\%$ of these surgical errors are preventable~\cite{zegers2011incidence}. Therefore, surgical data science (SDS)~\cite{maier2022surgical,maier2017surgical,vedula2017surgical} has emerged as a new discipline that aims to enhance the safety and quality of interventional healthcare through systematic modeling, analysis, and understanding of surgical data.

For this reason, important new tasks have been introduced, such as surgical workflow recognition~\cite{blum2008modeling,ding2025mosformer,padoy2012statistical,twinanda2016endonet,weede2012workflow}, surgical video segmentation~\cite{alapatt2021temporally,allan20192017,ni2022surginet,yang2022tmf}, and activity triplet recognition~
\cite{liu2024surgical,nwoye2022rendezvous,sharma2023surgical}. Existing methods typically approach these tasks using vision-only supervision at the categorical level. This creates a major bottleneck, since generating reliable labels requires detailed clinical expertise and extensive manual annotation by domain specialists, making large-scale data curation both time-consuming and resource-intensive.

A potential direction for advancing automatic tools for surgical understanding is large-scale foundation models. In particular, since the introduction of CLIP~\cite{radford2021learning}, multimodal representation learning has emerged as a promising approach for learning visual concepts through natural language supervision. 
With the growing availability of video-text pairs, developing foundation models capable of advanced vision-language understanding has become increasingly important. These pretrained models support a variety of downstream tasks, such as text-based retrieval~\cite{cheng2023vindlu,fu2021violet, lei2021less,li2023lavender,wang2023all,wang2022omnivl,wang2022internvideo}, video classification~\cite{li2023lavender,wang2023all,wang2022omnivl,wang2022internvideo}, and video captioning~\cite{li2023lavender,wang2023all,wang2022omnivl} in zero-shot, few-shot, and transfer-learning settings, obtaining performances that are on par with task-specific fully-supervised baselines. Their ability to excel even under limited supervision makes them especially promising for surgical data science, where labeled data is inherently scarce.

However, significant domain shifts between general-purpose vision-language models and the unique characteristics of surgical data, both in visual appearance and linguistic context, pose major challenges to their direct application, thereby resulting in suboptimal performance~\cite{yuan2023learning}. To address these challenges, a few recent studies have laid the foundation for surgery-specific pretraining frameworks that leverage video–text pairs from surgical lecture videos~\cite{honarmand2024vidlpro,yuan2024hecvl,yuan2024procedure,yuan2023learning}. Three key datasets driving this progress are the Surgical Video Lecture (SVL) dataset~\cite{yuan2023learning}, GenSurg+~\cite{honarmand2024vidlpro}, an extension of GenSurgery~\cite{schmidgall2024general}, and very recently SurgLaVi~\cite{perez2026surglavi} using both videos from YouTube and private recordings. In these datasets, textual annotations are automatically generated using automatic speech recognition (ASR) and further refined or expanded via large language models to ensure linguistic richness and accuracy. Existing methods can be broadly categorized into two approaches: the first~\cite{yuan2024hecvl,yuan2024procedure,yuan2023learning} encodes local video clips by extracting frame-level features using an Image Encoder followed by feature averaging; the second~\cite{honarmand2024vidlpro,perez2026surglavi} employs a Vision Transformer (ViT)-based Video Encoder with temporal attention to capture dynamic information within the local video clips. Furthermore, recent works~\cite{yuan2024hecvl,yuan2024procedure,perez2026surglavi} have augmented the SVL or SurgLaVi dataset with hierarchical procedural annotations, which serve as additional supervisory signals to enhance models’ understanding of surgical procedures. Neither of these two major modeling approaches attempts to capture long-term dependencies across the entire surgical scene. Capturing long-term temporal dependencies is crucial for developing a comprehensive understanding of surgical procedures. This contrasts sharply with common general computer vision tasks, where videos are typically very short, for instance, in the Something-Something V1 dataset~\cite{goyal2017something}, the average action and video duration is only $4.03$ seconds. In comparison, surgical procedures like, e.g., in the MultiBypass140 dataset~\cite{lavanchy2024challenges} are substantially longer, with an average video duration of approximately $91$ minutes ($\sim$1,355× longer), average phase duration of about $542$ seconds ($\sim$134× longer), and even finer action units such as steps averaging $142$ seconds ($\sim$35× longer).

Nevertheless, existing surgical vision-language works have achieved encouraging progress, but in contrast to the general computer vision domain, surgical vision-language approaches still lag behind task-specific fully-supervised approaches. The main reasons for this observation lie in the unique challenges posed by surgical videos and their inherent complexity. Unlike general computer vision datasets, where videos are typically short, actions are visually very distinct, and scenes are diverse, surgical procedures can span several hours and involve intricate, fine-grained actions within a constrained field of view, often occurring in the same operating space with minimal visual variation. Subtle tool motions, occlusions, and lighting variations further complicate surgical visual understanding. Furthermore, general vision-language models benefit from massive datasets such as WebVid10M~\cite{bain2021frozen} and HowTo100M~\cite{miech2019howto100m}; surgical vision-language datasets, however, remain orders of magnitude smaller. For instance, WebVid10M contains over $52,000$ hours of video, roughly $173$ times larger than the SVL dataset, which includes just $301$ hours of surgical lecture videos. In order to solve these unique challenges, new innovations are required to achieve finer-grained surgical domain understanding in data-sparse environments.

This work introduces \textbf{CliPPER}, a novel surgical video-text pretraining framework standing for \textbf{C}ontextua\textbf{l} V\textbf{i}deo-Language \textbf{P}retraining on Long-form Intraoperative Surgical \textbf{P}rocedures for \textbf{E}vent \textbf{R}ecognition. CliPPER is designed to enhance temporal understanding in surgical procedure videos, with a particular focus on contextual modeling of surgical procedures. As a first contribution, we curate and release a new public pretraining dataset constructed from $2,667$ YouTube videos. In total, the dataset comprises approximately $422$ hours of surgical video content and enables the construction of $41,182$ short video clip–text pairs for pretraining.
Unlike previous approaches that rely on Image-based Encoders, we adopt a Video Encoder architecture that builds on top of VindLU~\cite{cheng2023vindlu}, enabling richer modeling of temporal relations within individual short video clips. Furthermore, we capture temporal dynamics across clips from the same surgical video by introducing two novel temporal context modules, one for video and one for text, specifically designed to model higher-level temporal dependencies spanning hour-long procedures. Those modules enable the integration of contextual information beyond individual clips, allowing the model to better understand procedural flow and temporal coherence within complex surgical activities. 

However, learning solely from standard local video–text alignment objectives is insufficient to capture the broader procedural context. We therefore propose a complementary set of new pretraining strategies that explicitly leverage the temporal structure of long-form surgical videos. Among these, we introduce Contextual Video–Text Contrastive Learning, which aligns local video clips and their associated textual descriptions while conditioning on extended procedure-level context. Complementing this objective, we further incorporate a Cycle-Consistency Alignment constraint that enforces cross-temporal coherence and mitigates ambiguous or inconsistent clip–text correspondences in long-form surgical videos, particularly in surgical workflows where visually similar phases recur over extended time spans, leading to inherently many-to-one alignment ambiguities. 
Furthermore, contrastive alignment objectives are largely permutation-invariant and do not explicitly enforce temporal directionality within procedures. To address this limitation, we introduce Clip Order Prediction (COP), which trains the model to recover the correct temporal ordering of clips and their associated textual descriptions within the same surgical video. By providing an explicit order-sensitive supervisory signal, COP encourages the model to capture procedural progression and causal structure—properties that are critical in surgical workflows where the same visual patterns may occur at multiple stages but with different temporal semantics.
Finally, Frame–Text Matching (FTM) provides fine-grained supervision at the frame level. Whereas the prior objectives primarily encourage similar representations across all frames within the same clip, FTM explicitly aligns individual frame embeddings with their corresponding textual tokens. This ensures that each frame captures meaningful, task-relevant information, which is important to robustness for varying clip sizes and particularly for long clips, where subtle local differences can indicate semantically critical events. By enforcing high-information frame-level representations, FTM complements clip-level supervision and enables temporally precise visual–language modeling in long-form surgical videos.

We demonstrate the effectiveness of our method for workflow recognition at a larger scale than prior works by evaluating it on four public datasets covering four distinct surgical procedures. We also assess its performance on multiple public surgical benchmarks for finer-grained complex downstream surgical tasks, including step recognition, activity-triplet recognition, and instrument recognition. Across all these complex downstream tasks, our method achieves substantial performance improvements over existing approaches. 

%% file: sec/2_related.tex
\section{Related Work}
\label{sec:related}

\paragraph{Video-Language Pretraining}

Building foundation models for multi-modal representation learning with video-text pairs is gaining significant attention, largely due to their strong transferability and impressive zero-shot performance. Inspired by the Vision-Text Contrastive Learning introduced by CLIP~\cite{radford2021learning}, modifications are generated to extend it to the video domain called Video-Text Contrastive Learning~\cite{li2020learning,wang2022internvideo, xu2021videoclip}. Other works added additional pretraining losses such as Masked Video Modeling (MVM)~\cite{cheng2023vindlu,fu2021violet,tong2022videomae,wang2023videomae}, Next Token Prediction (NTP)~\cite{alayrac2022flamingo,sun2023emu}, Video-Text Matching (VTM)~\cite{cheng2023vindlu, fu2021violet,lei2021less, li2023lavender}, and Masked Language Modeling (MLM)~\cite{cheng2023vindlu,fu2021violet,li2023lavender,su2019vl} in order to strengthen modality fusion and modality specific self-supervision. Another line of work explores multiple training stages~\cite{li2023unmasked,wang2024internvideo2,DBLP:conf/icml/0003GYZYSFQW0HS24} to improve video representation. For instance, UMT~\cite{li2023unmasked} first pretrains using student-teacher distillation on images only, before switching to contrastive learning. InternVideo2~\cite{wang2024internvideo2} starts with unmasked video token prediction followed by multi-modal pretraining, while VideoPrism~\cite{DBLP:conf/icml/0003GYZYSFQW0HS24} reverses the process, applying contrastive learning before performing both global and local token distillation.

\paragraph{Video Temporal Modeling}
Understanding long-form videos is computationally demanding due to the high memory and processing requirements of modeling extended temporal sequences. Traditional approaches mitigate these challenges by using pre-computed features~\cite{abu2016youtube,donahue2015long,girdhar2017actionvlad,wu2021towards,yue2015beyond}, reducing frame rates~\cite{fu2021violet,korbar2019scsampler,lei2021less,lin2019tsm,wu2018compressed}, or employing efficient attention mechanisms, such as Structured State Space models (S4)~\cite{gu2021efficiently,islam2022long}, feature banks~\cite{wu2019long}, and memory caches~\cite{wu2022memvit}. Dual-stream models like SlowFast~\cite{feichtenhofer2019slowfast} and temporal modeling techniques, including spatial-temporal operations~\cite{bertasius2021space,wang2022deformable} or feature rolling~\cite{wang2023all}, further enhance efficiency. Despite these advancements, many video-language models still train on short clips (e.g., 4–64 frames)~\cite{ashutosh2023hiervl,cheng2023vindlu,fu2021violet,li2023lavender,xu2021videoclip}, which limits their ability to handle long-form videos. To overcome this, hierarchical models like HierVL~\cite{ashutosh2023hiervl} aggregate clip-level features to form long-form representations, while models like TemPVL~\cite{ma2023temporal} extend training to larger video sequences, incorporating fine-grained temporal boundary localization tasks to improve long-term understanding.

\paragraph{Surgical Video-Language Pretraining}

Recently, video-language pretraining has also been applied to multi-modal intraoperative surgical data~\cite{honarmand2024vidlpro, yuan2024hecvl,yuan2024procedure,yuan2023learning}. However, the scale of multi-modal surgical data clearly lags behind the general computer vision datasets.
Therefore, recent methods~\cite{yuan2024hecvl,yuan2024procedure} enhance data efficiency by incorporating hierarchical annotations to improve long-form procedural understanding. These approaches also employ feature aggregation techniques to represent higher-level hierarchical annotations, following a strategy similar to that used in HierVL~\cite{ashutosh2023hiervl}. In VidLPRO~\cite{honarmand2024vidlpro}, a new public pretraining dataset, GenSurg+, an extension of GenSurgery~\cite{schmidgall2024general}, is introduced alongside the adoption of a Video Encoder for local clip representation based on a ViT architecture.

%% file: sec/3_method.tex
\section{Method}
\label{sec:method} 

\begin{figure*}
    \centering
    \includegraphics[width=0.95\linewidth]{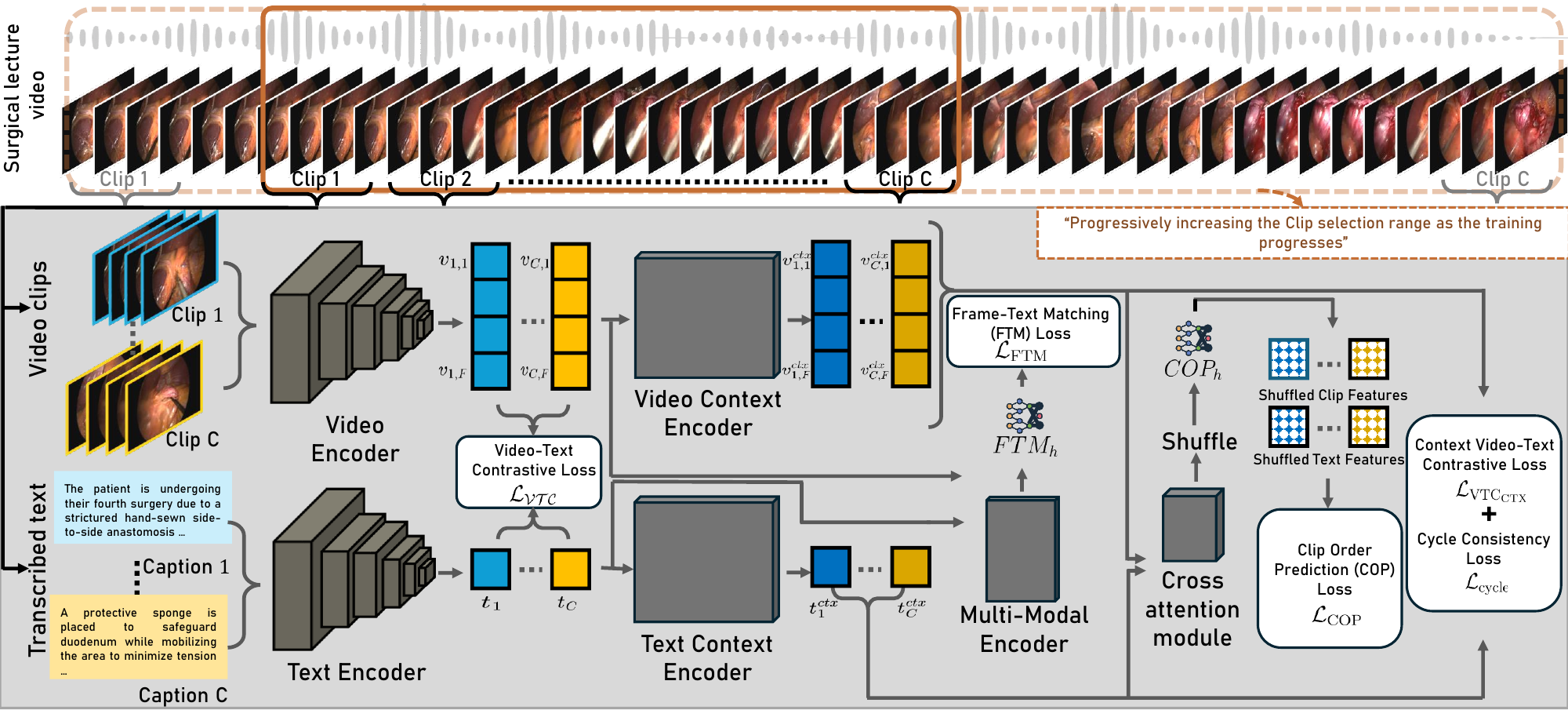}
    \caption{\textbf{CliPPER overview:} CliPPER processes multiple clips from the same video independently through modality-specific encoders with a Video Encoder for visual input and a Text Encoder for corresponding captions. The resulting pooled frame-level and text CLS embeddings are then passed through context encoders, which operate across clips from the same video to generate context-aware embeddings for each modality independently. On the Dual Encoder representations, we apply the standard video-text contrastive loss ($\mathcal{L_{VTC}}$). The context-aware representations are optimized through $\mathbf{VTC_{CTX}}$, a context-aware contrastive objective, together with a Cycle-Consistency Alignment loss that enforces bidirectional alignment consistency. In parallel, the model fuses all frame and text embeddings across clips using a Multi-Modal Encoder, enabling an additional objective: $\mathbf{FTM}$, which guides the model toward fine-grained alignment by learning to localize which frames from multiple clips from the same video match a given text. Lastly, we fuse the visual and textual contextual embeddings in a separate step as well by applying a single Cross-Attention layer. On top of these fused embeddings, we independently predict the temporal order of the elements called \textbf{COP}, enabling the model to reason about sequence information.}
    \label{fig:Overview}
\end{figure*}

We now present CliPPER, a novel method that builds on contrastive learning to improve contextual understanding and fine-grained video-text alignment in long-form surgical videos. The model is trained on both the surgical lecture videos from the SVL dataset~\cite{yuan2023learning} and the newly introduced YouTube videos. CliPPER is specifically designed to enhance both global context modeling and local semantic precision. 
\textbf{Contextual Video–Text Contrastive Learning} aligns local clips with their textual descriptions while conditioning on extended procedure-level context, addressing the limitations of standard local alignment.
\textbf{Cycle-Consistency Alignment} loss reduces many-to-one ambiguities that arise when visually similar phases recur over extended procedures.
\textbf{Clip Order Prediction (COP)} captures temporal dependencies by learning the correct sequential arrangement of clips and associated descriptions, enforcing order-sensitive procedural reasoning.
\textbf{Frame–Text Matching (FTM)} enforces fine-grained alignment between individual video frames and textual tokens, ensuring high-information frame embeddings and robustness to varying clip lengths.
For representation, we adopt a Dual-Encoder design with a BEiT-based Vision Transformer~\cite{bao2021beit} as the Video Encoder (with temporal attention) and BERT$_{base}$~\cite{devlin2019bert} as the Text Encoder. For multi-modal fusion, we integrate visual and textual features using the final layers of BERT as a Multi-Modal Encoder.

\subsection{Context Video-Text Contrastive Learning}

To capture the long-range dependencies crucial for understanding surgical procedures, we introduce a novel context-aware contrastive learning objective. Specifically, we sample a batch $B$ of videos. For each video, we randomly select a starting video clip with an index $i$, which serves as an anchor, and then sample an additional $C-1$ distinct video clips. These additional clips are chosen based on their temporal distance from the anchor clip with an index $i$. To progressively increase the contextual range, we employ a linearly growing schedule that expands the allowed distance of sampled clips from the same video, from roughly $15$ minutes in the first epoch to no temporal limitation in the final epochs. We construct the video-text pairs from the video clips, and their corresponding textual descriptions are generated from the audio tracks using Whisper V3~\cite{radford2023robust} partitioned into 45 second long clips and then rewritten using GPT-4.1~\cite{openai2025gpt4_1} thus yielding 41,182 video-text pairs for YouTube and 31,107 pairs for the SVL dataset. As illustrated in Figure~\ref{fig:Overview}, each video clip–text pair is independently processed through a Dual Encoder setup, comprising a Video Encoder and a Text Encoder, to obtain initial feature representations.

The Video Encoder generates embeddings for all patches across the $F$ frames of a given clip. We then apply average pooling over the patches within each frame with an index $f$ to obtain frame-level embeddings, denoted as $v_{i,f}$. These embeddings are subsequently concatenated across all $C$ clips of a video. Similarly, the Text Encoder outputs the CLS embeddings $t_i$ for each textual description of a given clip, which are also concatenated across all corresponding descriptions. These sequences are then passed through dedicated Video Context Encoder and Text Context Encoder, both implemented as Transformer Encoders~\cite{vaswani2017attention}, leveraging self-attention to model temporal and contextual dependencies within each modality separately. We apply relative positional encodings~\cite{dai2019transformer} to the frame embeddings before passing them into the Video Context Encoder, but we do not apply them to the text embeddings. This is because, during downstream evaluation, the textual prompts do not require explicit positional relationships; rather, a basic contextual understanding between different prompts is sufficient.

The resulting outputs are context-enhanced representations $v_{i, f}^{ctx}$ and $t_{i}^{ctx}$ for the $i$-th clip and text, as well as $f$-th frame of $F$ total frames. By computing the pairwise similarity between all video-text pairs, scaled by a temperature parameter $\sigma$ to control the sharpness of the distribution, in the batch $B$, we define the contextual video-text contrastive loss $\mathcal{L_{VTC_{CTX}}}$, as formalized in Equation~\ref{eq:CTX}:

\begin{equation}
\label{eq:CTX}
\resizebox{0.9\linewidth}{!}{$
\mathcal{L}_{\mathrm{VTC_{CTX}}} =
- \frac{1}{B \cdot C} 
\sum_{b=1}^{B} \sum_{i=1}^{C} 
\log 
\frac{
    \exp \big( {\frac{1}{F\cdot\sigma}\sum_{f=1}^{F}{(v_{b,i,f}^{ctx})}^{\top} t_{b,i}^{ctx}}\big)
}{
    \sum_{j=1}^{B} \sum_{k=1}^{C} \exp \big( \frac{1}{F\cdot\sigma}\sum_{f=1}^{F}{(v_{b,i,f}^{ctx})}^{\top} t_{j,k}^{ctx}\big)
}
$}.
\end{equation}

\subsection{Context Cycle-Consistency Alignment}

To enforce Cycle-Consistency between visual and textual contexts, we adopt the standard contrastive assumption that each clip in a batch has exactly one corresponding textual description (one-to-one matching within each video). This assumption serves as a training-time approximation to encourage discriminative cross-modal alignment. We begin by computing the similarity matrices $\mathbf{S^{v2t}}$, $\mathbf{S^{t2v}}$ between the two modalities as shown in Equation~\ref{eq:sim} and~\ref{eq:sim_matrix}:

\begin{equation}
\label{eq:sim}
    s^{v2t}_{i,j} = \big( \frac{1}{F\cdot\sigma}\sum_{f=1}^{F}{(v_{i,f}^{ctx})}^{\top} t_{j}^{ctx}\big), \\
    s^{v2t}_{i,j} = s^{t2v}_{j,i},
\end{equation}
    
\begin{equation}
\label{eq:sim_matrix}
\resizebox{0.88\linewidth}{!}{$
    \mathbf{S}^{v2t} = \{ s_{i,j}^{v2t} \mid i, j \in \{1,...,C\}\}, \\
    \mathbf{S}^{t2v} = \{ s_{i,j}^{t2v} \mid i, j \in \{1,...,C\}\} \in \mathbb{R}^{C \times C}
$},
\end{equation}
where $s_{i,j}^{v2t}$ and $s_{j,i}^{t2v}$ indicate the similarity score between the aggregated frame feature embeddings of clip $i$ and the text embedding of clip $j$. The similarity matrices are converted into probability distributions via a row-wise softmax, as shown in Equation~\ref{eq:prob}:
\begin{equation}
\label{eq:prob}
\mathbf{P}^{v2t} = \mathrm{Softmax}\!\left( \mathbf{S}^{v2t} \right),
\mathbf{P}^{t2v} = \mathrm{Softmax}\!\left(\mathbf{S}^{t2v} \right).
\end{equation}
To measure cycle-consistency, we compute the round-trip probability matrix, shown below in Equation~\ref{eq:cycle_prob}:
\begin{equation}
\label{eq:cycle_prob}
\mathbf{C} = \mathbf{P}^{v2t} \mathbf{P}^{t2v},
\mathbf{I} \in \mathbb{R}^{C \times C},
\end{equation}
which ideally approximates the identity matrix 
$\mathbf{I}$, since correct correspondences lie along the diagonal under the one-to-one matching assumption. Finally, the Cycle-Consistency Alignment loss is defined as the average squared Frobenius norm between $\mathbf{C}$ and $\mathbf{I}$ for a batch $B$ as shown in Equation~\ref{eq:cycle_loss}:
\begin{equation}
\label{eq:cycle_loss}
\mathcal{L}_{\text{cycle}} = 
\frac{1}{B} \sum_{b=1}^{B} 
\left\| \mathbf{C}_{b} - \mathbf{I} \right\|_2^2
\end{equation}
This objective encourages invertible cross-modal correspondences, ensuring that mappings from visual to textual context and back return to the original visual instance, thereby reducing many-to-one alignment ambiguities.

\subsection{Modality Fusion}

To enhance the discriminative power of video-language feature representations, we integrate information from both modalities using a Multi-Modal Encoder (MME). To further leverage the extended temporal context available in long-form surgical videos, we introduce two novel temporal supervision signals: FTM and COP.

The FTM objective targets fine-grained alignment at the frame level, as opposed to coarse video-level alignment in VTM, while still incorporating broader video context. To this end, we concatenate all frame embeddings $v_{i,f}$ from the Dual Encoders across C clips from the same video, along with a text token embedding $t_{i}$ from one of the clips as described in Equation~\ref{eq:ftm_pred}, where operator $[...]$ means concatenation of frame features.

\begin{equation}
\label{eq:ftm_pred}
\resizebox{0.9\linewidth}{!}{$
    \hat{m}_{i} = FTM_h(MME([v_{1,1}, ...,v_{1,F}, ..., v_{C,1}, ..., v_{C,F}], t_{i})), \\
    \hat{m}_{i} \in \mathbb{R}^{C \cdot F}.
$}
\end{equation}

These sequences are jointly processed by the MME. The MME consists of alternating layers of self-and cross-attention modules. During the cross-attention, the concatenated frame embeddings are the key and value vectors, while the text embedding serves as the query. The model then performs a binary classification for each frame, determining whether it aligns with the given text. This is achieved by feeding the fused frame features through a two-layer MLP (FTM$_h$). This step is repeated for each available text description within the video. Since the operating scene within a surgical video remains visually similar across most timestamps, applying this fine-grained loss effectively performs hard negative mining for each text description, as the other $C-1$ clip frame features lie initially close to each other in the embedding space, but, for fine-grained tasks, should be distinguishable. Since this is a binary classification task performed on each frame feature we use Binary Cross-Entropy (BCE) as shown in Equation~\ref{eq:ftm_loss}, where $m_{b,i}$ is a binary label vector of shape $\mathbb{R}^{C \cdot F}$ for a given batch sample $b$ and for text caption $i$.

\begin{equation}
\label{eq:ftm_loss}
    \mathcal{L}_{FTM} = \frac{1}{B\cdot C}\sum_{b=1}^{B}\sum_{i=1}^{C}BCE(\hat{m}_{b,i}, {m_{b,i})}
\end{equation}

For Clip Order Prediction (COP) objective, we feed the contextualized frame embeddings $v^{ctx}_{i,f}$ and text embeddings $t_{i}^{ctx}$ into a single Cross-Attention layer to fuse the contextualized modalities. Afterwards, the fused frame embeddings of embedding dimension $D$ for each clip are concatenated to form a single clip representation embedding $v_{i}^{ctx} \in \mathbb{R}^{F \cdot D}$. The clip and text embeddings are then randomly shuffled within each video and passed through a two-layer MLP (COP$_h$) to predict the original temporal order of the clips and text descriptions. We treat this as a multi-class classification problem and employ a Categorical Cross-Entropy loss function. This objective encourages the model to develop a stronger understanding of the temporal structure and the progression of events throughout the video.

%% file: sec/4_experiments.tex
\section{Experiments}
\label{experiments}

\paragraph{Pretraining and Downstream Datasets:}

We use public YouTube videos for pretraining and denote this setting as \textbf{“YT”} in the tables. In addition, we also pretrain on the surgical lecture videos from the SVL dataset, obtained with permission from the authors of SurgVLP~\cite{yuan2023learning}, for which we generate transcripts using our own audio transcription pipeline (identical steps as in YT pipeline); this setting is denoted as \textbf{“SVL”}. When pretraining is performed on both datasets jointly, we report the setting as \textbf{“All”}.

For all the downstream evaluation, we use publicly available datasets. We perform zero-shot surgical workflow phase recognition on $4$ publicly available datasets, covering distinct surgical procedures. The public datasets include Cholec80~\cite{twinanda2016endonet} (cholecystectomy), AutoLaparo~\cite{wang2022autolaparo} (hysterectomy), MultiBypass140~\cite{lavanchy2024challenges} (gastric bypass), and GraSP~\cite{ayobi2025pixel} (prostatectomy). Phase recognition is formulated as a clip-wise classification task, with $7$ phase labels in both Cholec80 and AutoLaparo, $13$ in MultiBypass140, and $11$ in GraSP.
The numbers reported in the following tables are computed on the official test splits of each dataset. In addition, we investigate finer-grained aspects of surgical workflow by performing step recognition on MultiBypass140~\cite{lavanchy2024challenges} with $46$ unique labels, and on GraSP~\cite{ayobi2025pixel} with $21$ labels.

An even finer-grained task is activity triplet recognition, which we evaluate in a zero-shot multi-label classification setting on CholecT50~\cite{nwoye2022rendezvous} (cholecystectomy) with $100$ unique labels and on ProstaTD~\cite{chen2025prostatd} (prostatectomy) with $89$ unique labels. In addition, we perform surgical instrument recognition on both datasets. In the surgical domain, activity triplets are defined as a combination of the instrument, the action it performs, and the target organ or tissue, i.e., [instrument, verb, target]. Similarly to the other tasks, we report the results on the test set, which specifically for CholecT50 refers to the official Rendezvous split~\cite{nwoye2022rendezvous}.

\begin{table*}
\centering
\caption{Zero-shot surgical phase recognition results: We report on F1-Score average across $4$ different prompts. Our approach outperforms other methods significantly across all datasets. 
VindLU baseline uses the original pretrained weights from its source repository and VindLU-SVL is trained on the same narrations as Ours-SVL. Ours-YT is on the public YouTube narrations, and Ours-All specifies both SVL and YouTube videos combined.
All other baselines are trained on the SVL as specified in the corresponding works.
}
\resizebox{0.78\linewidth}{!}{%
\begin{tabular}{lccccccl}\toprule
         Model  & Cholec80  & AutoLaparo    & StrasBypass70  & BernBypass70 & GraSP & Average\\\midrule
VindLU~\cite{cheng2023vindlu}    & $8.2$ & $7.7$  &  $2.5$   &   $2.6$  &   $2.8$  &  $4.7$ \\ \midrule
SurgVLP~\cite{yuan2023learning}  & $23.3$ & $10.8$ & $14.1$ &  $7.8$   & $7.6$ &  $12.7$\\
HecVL~\cite{yuan2024hecvl} & $22.5$ & $19.3$ & $21.8$  &  $15.4$ & $6.0$ & $17.0$\\
PeskaVLP~\cite{yuan2024procedure} & $30.5$ & $25.4$ & $26.5$ &  $19.2$   & $7.7$ & $21.8$ \\
\midrule
VindLU-SVL   & $29.3$ & $17.9$ &  $31.5$  &  $18.3$  & $14.9$ & $22.4$\\ \midrule
Ours-SVL  & $30.6$ & $31.8$ & $\underline{34.8}$  &  $\underline{21.7}$  & $16.9$  & $27.2$\\
Ours-YT  & $\underline{34.9}$ & $\mathbf{50.0}$ & $30.3$  &  $18.1$  & $\underline{33.3}$  & $\underline{33.3}$\\ \midrule
Ours-All  & $\mathbf{38.3}$ & $\underline{49.4}$ & $\mathbf{37.9}$  &  $\mathbf{24.1}$  & $\mathbf{34.1}$  & $\mathbf{36.8}$\\ \bottomrule
\end{tabular}
}
\label{Tab:Phase}
\end{table*}

\paragraph{Implementation Details:}
Our Video Encoder backbone is based on the BEiT model~\cite{bao2021beit}, augmented with an additional temporal attention module placed before the spatial attention within the image transformer. The Text and Multi-Modal Encoders are initialized using BERT$_{base}$~\cite{devlin2019bert}. Both the Text and Video Context Encoders are implemented as $4$-layer vanilla Transformer Encoders, each with $2$ attention heads. The temperature $\sigma$ is a learnable parameter as in CLIP~\cite{radford2021learning} and initialized with $0.07$. All loss functions are weighted equally.

We train the model on $2$ NVIDIA TESLA V100 GPU with a DRAM of $32$ GB for $20$ epochs using the AdamW optimizer~\cite{kingma2014adam,loshchilov2017decoupled} with a learning rate of $1 \times 10^{-4}$, following a cosine learning rate schedule~\cite{loshchilov2016sgdr}. During training, we sample $8$ clips per video with a batch size of $32$. Each clip consists of $4$ sparsely sampled frames, resized to $224$×$224$ pixels. Furthermore, during training, we alternate between GPT-generated and Whisper-generated captions with equal probability.

\paragraph{Evaluation Setup:} At inference time, the videos are sampled at $1$ FPS for both tasks. All frames within a clip are passed through the Video/Image Encoders, while all clips from a video are processed by the Contextual Encoders. Our final zero-shot predictions are obtained by averaging contextual \{$v_{i,f}^{ctx}$, $t_{i}^{ctx}$\} and non-contextual \{$v_{i,f}$, $t_{i}$\} alignment scores, unless otherwise specified. For surgical phase and step recognition, we use $4$-second clips to preserve temporal structure, in the supplementary material we also show results for different temporal windows. For triplet recognition, we perform single-frame prediction. Given the reliance of zero-shot evaluations on high-quality textual prompts, we utilize four distinct prompts for each label in the phase recognition task. This includes the original prompts used in PeskaVLP~\cite{yuan2024procedure}, along with three additional prompts generated using Gemini 2.0 Flash~\cite{google2025gemini2flash}, GPT-4o~\cite{hurst2024gpt}, and Claude 3.7 Sonnet~\cite{anthropic2025claude37sonnet}. For triplet and step recognition, we generate $4$ individual prompts using GPT-4.1, Claude 4.5 Sonnet~\cite{claude-sonnet-4.5}, Gemini 2.5 Flash~\cite{gemini-flash-2.5}, and DeepSeekV3. All the utilized prompts will be released with the source code. 
All reported results in the subsequent tables are averaged over the respective sets of prompts to ensure robustness and reduce prompt-specific bias. As baselines, we include all publicly available pretrained methods: SurgVLP~\cite{yuan2023learning}, HecVL~\cite{yuan2024hecvl}, PeskaVLP~\cite{yuan2024procedure}, as well as VindLU~\cite{cheng2023vindlu}, which we pretrain on the SVL videos.

\subsection{Zero-shot Surgical Workflow Recognition}
\label{Workflow}

The primary task for evaluating video–language models in SDS is \textit{workflow recognition}, as it reflects a model's ability to comprehend the overall surgical procedure. This task is typically evaluated at two levels: a coarser level known as \textit{phase recognition}, and a finer-grained level referred to as \textit{step recognition}. We use the F1-score as the metric for both tasks due to the inherent class imbalance in each of the used datasets. Unlike prior approaches~\cite{yuan2024hecvl,yuan2024procedure,yuan2023learning}, which perform evaluation at the individual frame level, we operate on short clips of 4 frames ($\sim$ 4 seconds). This choice aligns with our temporally aware video–language architecture and enables more faithful modeling of dynamic surgical activity. Additionally, we report results for different temporal windows in the supplementary material. Furthermore, the \textit{zero-shot} setting implies that neither the videos nor their corresponding labels from the evaluation datasets are seen by the model during training. The results presented in Table~\ref{Tab:Phase} indicate that compared to the best baseline (22.4\%), CliPPER achieves 27.2\% under the same-dataset pretraining setting (+4.8\% absolute / +21.4\% relative) and 36.8\% when pretrained on all data (+14.4\% absolute / +64.3\% relative). Pretraining on the YouTube corpus alone yields 33.3\%, corresponding to a +6.1\% absolute (+22.4\% relative) improvement over pretraining on SVL only.
A closer look reveals that YouTube-only pretraining underperforms SVL on gastric bypass but provides substantial gains for hysterectomy (AutoLaparo) and prostatectomy (GraSP). This trend is consistent with the underlying pretraining distribution, as SVL contains proportionally more gastric bypass videos but fewer hysterectomy and prostatectomy cases.
Combining both datasets leads to strong improvements across nearly all phase recognition benchmarks, with only a marginal decrease observed on AutoLaparo.

\begin{table}
\centering
\caption{Zero-shot surgical step recognition results: We report on F1-score averaged across $4$ different prompts.
 VindLU uses the original pretrained weights from the VindLU source repository and VindLU-SVL is trained on the same text and videos as Ours-SVL. Ours-YT is on the public YouTube narrations, and Ours-All specifies both SVL and YouTube videos combined. All other baselines are trained on the SVL dataset.}
\resizebox{0.80\linewidth}{!}{%
\begin{tabular}{lcccl}\toprule
         Model  & StrasBypass70  & BernBypass70 & GraSP & Average\\\midrule
VindLU~\cite{cheng2023vindlu}    & $0.3$ & $0.3$  &  $0.7$   &   $0.4$ \\ \midrule
SurgVLP~\cite{yuan2023learning}  & $4.5$ & $3.1$ & $7.6$  & $5.0$\\
HecVL~\cite{yuan2024hecvl} & $6.0$ & $5.0$ & $6.0$  & $5.7$\\
PeskaVLP~\cite{yuan2024procedure} & $5.2$ & $4.1$ & $7.7$  &  $5.7$\\
\midrule
VindLU-SVL & $11.7$ & $\underline{7.6}$ &  $6.3$   &  $8.7$\\\midrule
Ours-SVL  & $\underline{13.3}$ & $6.9$ & $6.8$  &  $9.0$\\
Ours-YT  & $11.6$ & $7.5$ & $\underline{16.6}$  &  $\underline{11.9}$\\ \midrule
Ours-All  & $\mathbf{15.8}$ & $\mathbf{9.3}$ & $\mathbf{16.8}$  &  $\mathbf{14.0}$\\ \bottomrule
\end{tabular}
}
\label{Tab:Step}
\end{table}

Similarly, for the finer-grained step recognition task (Table~\ref{Tab:Step}), CliPPER achieves an absolute improvement of 0.3\% when pretrained on the same dataset and 5.3\% when pretrained on all available data, corresponding to relative gains of 3.5\% and 61\%, respectively. These results demonstrate that CliPPER significantly enhances surgical workflow understanding, even without any task-specific labeled data. In contrast, the general computer vision pretrained model VindLU performs substantially worse, highlighting that directly applying standard CV models to SDS tasks remains challenging.

\subsection{Zero-shot Activity Triplet Recognition}
\label{Triplet}

\begin{table}
\centering
\caption{Zero-shot surgical triplet (\textit{ivt}) and instrument recognition results (\textit{i}). We report the mean Average Precision (mAP) averaged across four different prompts. Our approach consistently outperforms prior methods on both datasets.
}
\resizebox{0.85\linewidth}{!}{%
\begin{tabular}{lcccccc}
\toprule
& \multicolumn{2}{c}{CholecT50} & \multicolumn{2}{c}{ProstaTD} & \multicolumn{2}{c}{Average} \\
\cmidrule(lr){2-3} \cmidrule(lr){4-5} \cmidrule(lr){6-7}
 Model & $ivt_{mAP}$ & $i_{mAP}$ & $ivt_{mAP}$ & $i_{mAP}$ & $ivt_{mAP}$ & $i_{mAP}$ \\ 
\midrule
VindLU~\cite{cheng2023vindlu}  & $3.0$ & $26.4$ & $3.4$ & $36.7$ & $3.2$ & $31.5$\\
\midrule
SurgVLP~\cite{yuan2023learning} & $3.7$ & $27.5$ & $3.9$ & $39.1$ & $3.8$ & $33.3$\\
HecVL~\cite{yuan2024hecvl} & $3.5$ & $27.9$ & $4.2$ & $41.1$ & $3.9$ & $34.5$\\
PeskaVLP~\cite{yuan2024procedure} & $4.3$ & $32.2$ & $4.3$ & $42.6$ & $4.3$ & $37.4$ \\
\midrule
VindLU-SVL & $5.0$ & $36.3$ & $4.4$ & $40.4$ & $4.7$ & $38.4$ \\
\midrule
Ours-SVL & $5.2$ & $\underline{36.7}$ & $5.1$ & $41.1$ & $5.2$ & $38.9$\\
Ours-YT & $\underline{5.8}$ & $33.2$ & $\underline{7.4}$ & $\mathbf{48.6}$ & $\underline{6.6}$ & $\underline{40.9}$\\ \midrule
Ours-All & $\mathbf{6.6}$ & $\mathbf{40.7}$ & $\mathbf{7.1}$ & $\underline{47.0}$ & $\mathbf{6.9}$ & $\mathbf{43.9}$\\
\bottomrule
\end{tabular}%
}
\label{Tab:Triplet}
\end{table}

\begin{table*}[t!]
\centering
\caption{Ablation studies on phase recognition: We evaluate the performance of various model configurations using the F1-score. We compute the F1-score for each video and compute the average F1-score across all videos, then further average over four distinct text prompts.
When the context-aware contrastive objective ($VTC_{CTX}$ here in short CTX) is not used, the model defaults to using only the base frame ($v_{i,f}$) and text ($t_{i}$) embeddings for alignment. When the Video Encoder (VE) is not selected, then the ResNet50 image encoder is used.}
\resizebox{0.80\linewidth}{!}{%
\begin{tabular}{lcccccccccl}\toprule
         COP & CTX & FTM & CYC & VE & Cholec80 & AutoLaparo & StrasBypass70 & BernBypass70 & GraSP & Average \\\midrule
$\times$ & $\times$ & $\times$ & $\times$ &  $\times$  & $29.4$ & $18.0$ & $30.4$ & $17.4$ &  $14.9$ & $22.4$ \\\midrule
$\times$ & $\checkmark$ & $\checkmark$ & $\checkmark$ & $\checkmark$ & $\underline{31.4}$ & $26.5$ & $33.9$ & $\underline{21.6}$ & $15.6$ & $25.8$\\
$\checkmark$ & $\checkmark$ & $\times$ & $\checkmark$ & $\checkmark$ & $27.9$ & $\underline{30.5}$ & $\underline{34.2}$ & $20.1$ & $\underline{16.2}$ & $25.8$ \\
$\checkmark$ & $\checkmark$ & $\checkmark$ & $\times$ & $\checkmark$ & $\mathbf{32.3}$ & $29.7$ & $34.1$ & $21.2$ & $\underline{16.8}$ & $\underline{26.8}$ \\
$\checkmark$ & $\checkmark$ & $\checkmark$ & $\checkmark$ & $\checkmark$  & $30.6$ & $\mathbf{31.8}$ & $\mathbf{34.8}$ & $\mathbf{21.7}$ & $\mathbf{16.9}$ & $\mathbf{27.2}$\\\midrule
$\times$ & $\times$ & $\times$ & $\times$ & $\times$  & $18.3$ & $16.7$ & $30.1$ & $16.2$ & $10.4$ & $18.3$\\
$\checkmark$ & $\checkmark$ & $\checkmark$ & $\checkmark$ & $\times$   & $27.5$ & $27.2$ & $30.6$ & $18.4$ & $14.7$ & $23.7$\\\bottomrule
\end{tabular}
}
\label{Tab:Ablation}
\end{table*}

Table~\ref{Tab:Triplet} reports zero-shot performance on \textit{CholecT50} and \textit{ProstaTD} for surgical triplet and instrument recognition. Using VindLU-SVL as the strongest baseline, \textbf{Ours-SVL} improves performance from 5.0\% to 5.2\% for triplet recognition and from 38.4\% to 38.9\% for instrument recognition, corresponding to relative gains of $\mathbf{+4.0\%}$ and $\mathbf{+1.3\%}$, respectively. Scaling pretraining to all available data further boosts performance to 6.6\% and 43.9\%, yielding substantial relative improvements of $\mathbf{+32.0\%}$ for triplet recognition and $\mathbf{+14.3\%}$ for instrument recognition over VindLU-SVL. These results highlight the benefit of explicitly modeling long-form surgical context for zero-shot surgical understanding.

It is worth noting that surgical triplet recognition represents an especially challenging task, as it requires understanding complex spatial and temporal relationships between surgical instruments, actions, and anatomical targets. The clear improvements achieved by \textbf{CliPPER} highlight the effectiveness of our finer-grained pretraining tasks as well as contextual modeling of the video in capturing these higher-order semantic dependencies, resulting in more structured and context-aware surgical video representations.

\subsection{Ablation Studies}
Table~\ref{Tab:Ablation} analyzes the contribution of the proposed pretraining objectives as well as the impact of using a temporal Video Encoder instead of an image-based backbone. Incorporating the temporal objectives, COP, $VTC_{CTX}$, and Cycle-Consistency Alignment loss, consistently improves visual–temporal alignment and leads to better workflow understanding across surgical procedures, with the largest gains observed on long-form scenarios such as AutoLaparo.

Adding the fine-grained FTM objective yields further improvements, particularly for procedures with subtle visual differences (e.g., GraSP and Bypass), indicating the benefit of sub-clip supervision for disambiguating visually similar phases.

Overall, combining all proposed objectives results in an average gain of $\mathbf{4.8\%}$ over the strengthened video–language baseline VindLU. Replacing the temporal Video Encoder with a standard ResNet50 leads to the expected performance drop, yet the model still outperforms the temporal baseline trained without our objectives. Moreover, when compared to a ResNet50 model trained with a standard CLIP-style VTC loss, our objectives provide a consistent $\mathbf{5.4\%}$ improvement.

These results demonstrate that the proposed training strategy is robust and largely encoder-agnostic, making it applicable across both image- and video-based backbone architectures.

%% file: sec/5_conclusion.tex
\section{Conclusion}
\label{sec:conclusion}

We introduced CliPPER, a novel video-language pretraining framework tailored for the intraoperative surgical domain, with a focus on capturing temporal context and enabling fine-grained video-text alignment. The framework incorporates four newly proposed objectives: Contextual Video-Text Contrastive Learning, Clip Order Prediction, Cycle-Consistency Alignment loss, and Frame-Text Matching, to enhance temporal and semantic understanding across surgical videos. CliPPER achieves state-of-the-art zero-shot performance on multiple public surgical workflow recognition benchmarks, with overall improvements over prior SOTA of \textbf{14.4\%} and \textbf{8.3\%} in phase and step recognition, respectively. 
Additionally, it outperforms prior works in both triplet and instrument recognition, achieving overall mAP gains of \textbf{2.6\%} and \textbf{6.5\%}, respectively, demonstrating the effectiveness of the proposed approach.

%% file: sec/6_acknowledgement.tex
\section{Acknowledgement}

We would like to express our sincere appreciation to educational platforms such as WebSurg (IRCAD), EAES, and YouTube for their commitment to providing high-quality educational content that is freely accessible to learners worldwide. We are particularly grateful to the clinicians who generously share their time and expertise to create and publish educational material on these platforms, thereby making this research possible.

This work was supported by French State Funds managed by the Agence Nationale de la Recherche (ANR) under Grant ANR-22-FAI1-0001 (project DAIOR) and Grant ANR-10-IAHU-02 (IHU Strasbourg).

%% file: sec/7_appendix.tex
\clearpage
\section{Appendix}

\subsection{Impact of Temporal Window Size on Workflow Recognition}

\begin{figure}[h]
    \centering
    \includegraphics[width=0.95\linewidth]{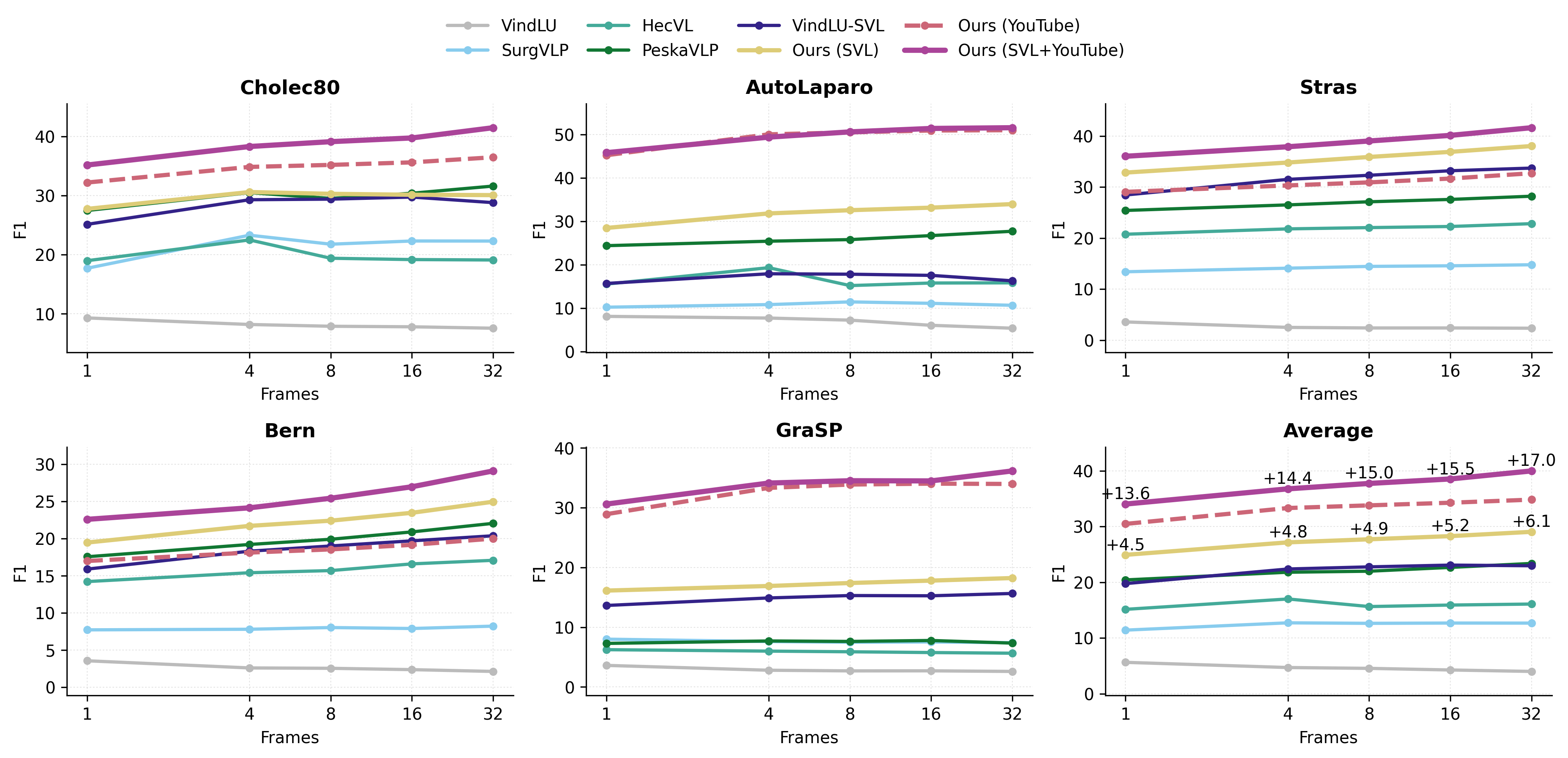}
    \caption{\textbf{Zero-shot Phase Recognition} at varying temporal windows: 
The final plot shows the average performance across all datasets. 
The highlighted values indicate the improvement over the strongest baseline for both \textbf{Ours-SVL} and \textbf{Ours (SVL + YouTube)}.}
    \label{fig:Temp_Phase}
\end{figure}

\begin{figure}[h]
    \centering
    \includegraphics[width=0.95\linewidth]{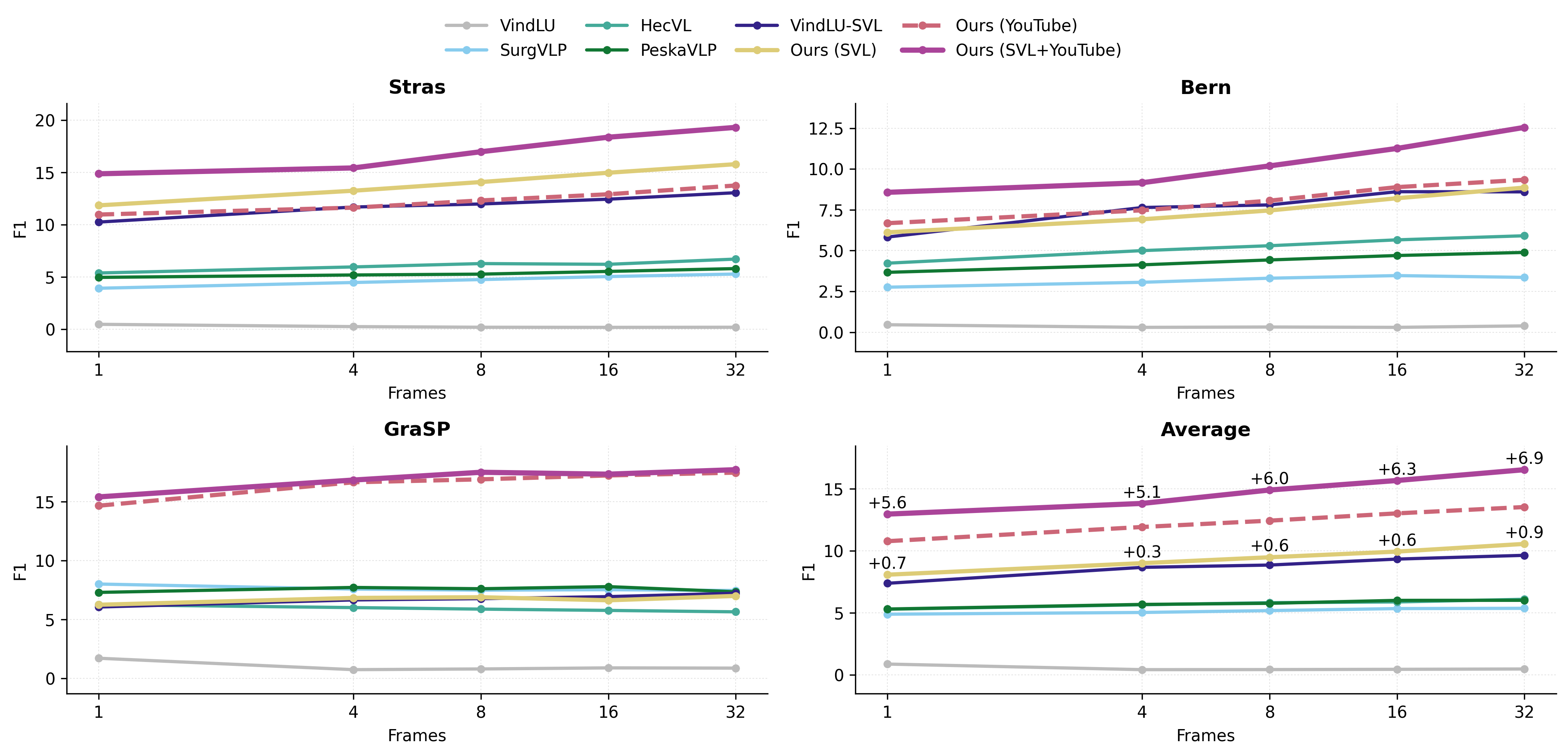}
    \caption{\textbf{Zero-shot Step Recognition} at varying temporal windows: 
The final plot shows the average performance across all datasets. 
The highlighted values indicate the improvement over the strongest baseline for both \textbf{Ours-SVL} and \textbf{Ours (SVL + YouTube)}.}
    \label{fig:Temp_Steps}
\end{figure}

Figures~\ref{fig:Temp_Phase} and~\ref{fig:Temp_Steps} present zero-shot phase and step recognition results for all baselines and our models across temporal clip windows ranging from 1 to 32 frames. The datasets are sampled at 1 FPS, i.e., 32 frames correspond to 32-second clips. For both tasks, the results consistently show that our method benefits more strongly from increasing temporal context than the baselines, highlighting the effectiveness of our contextual modeling for longer temporal windows. In contrast to our approach, most baselines operate on clip-level representations without explicit modeling of cross-clip temporal context, which likely limits their ability to benefit from larger temporal windows.

\subsection{Sampling Ablations}

Table~\ref{Tab:Ablations} presents additional ablation studies conducted on the SVL dataset and highlights the importance of progressive sampling, particularly for longer procedures such as hysterectomy (AutoLaparo). Furthermore, we observe that combining Whisper- and GPT-generated captions improves alignment performance compared to using only one source. This finding aligns with prior work suggesting that increased variability in textual descriptions improves robustness and reduces overfitting to specific captions.

\begin{table}[h]
\centering
\caption{\textbf{Sampling Ablations:} This table highlights the importance of progressive sampling and the benefit of combining Whisper- and GPT-generated captions during pretraining. All pretraining experiments are conducted on the SVL dataset.}
\resizebox{0.88\linewidth}{!}{%
\begin{tabular}{lcccccc}
\toprule
Model & Cholec80 & AutoLaparo & StrasBypass70 & BernBypass70 & GraSP & Average\\ \midrule
w/o progressive sampling & 31.0 & 21.9 & 33.1 & 20.3 & 15.8 & 24.4 \\
only Whisper & 28.1 & \textbf{32.4} & 29.0 & 19.1 & 13.5 & 24.4 \\
only GPT & \textbf{32.5} & 30.8 & 31.9 & 19.8 & 14.6 & 25.9 \\
\midrule
Ours-SVL & 30.6 & 31.8 & \textbf{34.8} & \textbf{21.7} & \textbf{16.9} & \textbf{27.2} \\
\bottomrule
\end{tabular}
}
\label{Tab:Ablations}
\end{table}